\documentclass[conference]{IEEEtran}
\IEEEoverridecommandlockouts
\IEEEpubid{
\begin{minipage}{\textwidth}\ \\[12pt]
\vspace*{1cm}
\\© 2025 IEEE. Personal use of this material is permitted. Permission from IEEE must be obtained for all other uses, in any current or future media, including reprinting/republishing this material for advertising or promotional purposes, creating new collective works, for resale or redistribution to servers or lists, or reuse of any copyrighted component of this work in other works.
\end{minipage}
}
\usepackage{graphicx}%
\usepackage{multirow}%
\usepackage{stfloats}
\usepackage{amsmath,amssymb,amsfonts}%
\usepackage{amsthm}%
\usepackage{mathrsfs}%
\usepackage{xcolor}%
\usepackage{textcomp}%
\usepackage{manyfoot}%
\usepackage{booktabs}%
\usepackage{algorithm}%
\usepackage{algorithmicx}%
\usepackage{algpseudocode}%
\usepackage{listings}%
\usepackage{makecell}
\usepackage{hyperref}
\usepackage{xspace}
\usepackage{float}

\newcommand{\eat}[1]{}

\newcommand{\eg}{e.\,g.,\xspace}

\def\BibTeX{{\rm B\kern-.05em{\sc i\kern-.025em b}\kern-.08em
    T\kern-.1667em\lower.7ex\hbox{E}\kern-.125emX}}

\begin{document}

\author{\IEEEauthorblockN{ Steffen Meinert\textsuperscript{*} and Philipp Schlinge\textsuperscript{*}}
\IEEEauthorblockA{\textit{Semantic Information Systems Group} \\
\textit{Osnabrück University}\\
Osnabrück, Germany \\
\{steffen.meinert, philipp.schlinge\}@uos.de}
\and
\IEEEauthorblockN{Nils Strodthoff}
\IEEEauthorblockA{\textit{AI4Health Division, Faculty VI} \\
\textit{Carl von Ossietzky Universität Oldenburg}\\
Oldenburg, Germany \\
nils.strodthoff@uol.de}
\and
\IEEEauthorblockN{Martin Atzmueller}
\IEEEauthorblockA{\textit{Semantic Information Systems Group} \\
\textit{Osnabrück University, and DFKI}\\
Osnabrück, Germany \\
martin.atzmueller@uos.de}
}
\title{ProtoMask: Segmentation-Guided Prototype Learning
\thanks{\textsuperscript{*}Steffen Meinert and Philipp Schlinge contributed equally to this work. This work was supported by the German Federal Ministry for Economic Affairs and Energy (BMWE), project \emph{FRED} (FKZ: 01MD22003E), and by the German Research Foundation (DFG), grant MODUS-II (316679917, AT 88/4-2).}
}

\maketitle

\begin{abstract}
XAI gained considerable importance in recent years. Methods based on prototypical case-based reasoning have shown a promising improvement in explainability. However, these methods typically rely on additional post-hoc saliency techniques to explain the semantics of learned prototypes. Multiple critiques have been raised about the reliability and quality of such techniques. For this reason, we study the use of prominent image segmentation foundation models to improve the truthfulness of the mapping between embedding and input space. We aim to restrict the computation area of the saliency map to a predefined semantic image patch to reduce the uncertainty of such visualizations. To perceive the information of an entire image, we use the bounding box from each generated segmentation mask to crop the image. Each mask results in an individual input in our novel model architecture named ProtoMask. We conduct experiments on three popular fine-grained classification datasets with a wide set of metrics, providing a detailed overview on explainability characteristics. The comparison with other popular models demonstrates competitive performance and unique explainability features of our model. \url{https://github.com/uos-sis/quanproto}
\end{abstract}

\begin{IEEEkeywords}
Interpretable AI, XAI, Prototype-based Models
\end{IEEEkeywords}

\section{Introduction}
Applying deep learning models
is often subject to constraints, particularly in high-risk areas where a model's decision can affect or endanger human lives. This has led to considerable advances in explainable deep learning, to make the decisions of deep neural networks more understandable to humans~\cite{minh2022explainable,AFKS:24}. 
Prototype-based models are a notable research direction, based on modern case-based reasoning approaches~\cite{chen2019looks}. A model learns a set of prototypical class representations within the embedding space, aiming to explain the classification process using a small number of prototypes.
%
However, the intrinsic explanation consists of a set of prototypes in the embedding space, making them difficult to interpret. Relying on post-hoc methods, which introduce errors in the mapping process from embedding to input space, is a major drawback of such models~\cite{hoffmann2021looks}.
In this paper, we address that problem, by improving the alignment between prototypes and object parts, focusing on informative image regions.

Approaches such as ProtoPNet~\cite{chen2019looks} utilized a simple upscaling technique, as pioneered by Grad-CAM~\cite{selvaraju2017grad}.
This method, which explains a prototype by highlighting important regions in real-world examples, was a simple step to facilitate the interpretation of prototypes. Many later studies adopted this approach~\cite{rymarczyk2022interpretable},~\cite{nauta2023pip},~\cite{donnelly2022deformable}.

Furthermore, recent research investigated the accuracy of saliency maps regarding prototype explanations~\cite{hoffmann2021looks}, revealing a discrepancy between the importance of saliency maps and the actual network behaviour. This motivates the use of more sophisticated saliency methods in order to improve the alignment between explanations and model behaviour~\cite{gautam2023looks}.  Nonetheless, current CNNs are often so deep that the receptive field of a prototype covers the entire image, leading to unhelpful  explanations as changes in unimportant regions for the saliency map influence the prototype regardless~\cite{xu2023sanity}. In contrast, alternative approaches such as FeatInv~\cite{neukirch2025featinv} train a generative model to learn a probabilistic mapping from feature space to input space and yields high-fidelity visualizations in input space, but still have the same problem operating on the information from the entire image.

To the best of the authors' knowledge, all recently introduced prototype models use the combination of large neural networks and simple post-hoc methods to generate saliency maps as a prototype representation. The general idea of restricting the receptive field of a prototype or the use of segmentation masks is not well studied in the context of prototype models. Works such as Sun et al.~\cite{sun2024explain} and Kong et al.~\cite{kong2024lce} studied the use of segmentation masks as concept explanations in a model agnostic design. Furthermore, works such as Kim et al.~\cite{kim2024vision} used segmentation masks to create more semantic image tokens in visual transformer models.

Altogether, this motivated us to propose \emph{ProtoMask}, our novel approach presented in this paper: ProtoMask is a prototype-based image classification model that operates on multiple views of the same image, created using segmentation masks to limit the prototype's receptive field and post-hoc saliency methods' operation area to a small image patch, respectively. This improves the alignment between prototypes and object parts, which increases the trustworthiness of visual explanations by restricting the error-prone post-hoc saliency method to an already informative image region.

\section{Related Work}
There has been a steady increase in research efforts on prototype-based networks in the general image classification domain. A major design choice is whether a prototype is distinct to a class or can represent a general feature for multiple classes. The distinct case was adopted by one of the earliest networks, ProtoPNet~\cite{chen2019looks}, and derivatives such as Deformable ProtoPNet~\cite{donnelly2022deformable} and ST-ProtoPNet~\cite{wang2023learning}. We have also opted to use this general approach, aiming to facilitate the interpretation of visual prototype representations. Prototype networks that learn shared prototypes include ProtoPShare~\cite{rymarczyk2021protopshare}, ProtoPool~\cite{rymarczyk2022interpretable} and PIPNet~\cite{nauta2023pip}. This enables the network to represent more general characteristics with a prototype, but can also increase the complexity of prototypes. The previously mentioned networks all use a linear classification design, including ours. An alternative is to use a hierarchical tree structure, as in HPNet~\cite{hase2019interpretable} and ProtoTree~\cite{nauta2021neural}. With the popularity of transformer models, there are also models that use visual transformer as feature extractors in contrast to the standard CNN backbone models such as VGG, ResNet and DenseNet. ProtoPFormer~\cite{xue2022protopformer} and ProtoViT~\cite{ma2024interpretable} use distinct prototypes in a linear classification process in combination with such transformer backbones. A tree-based approach with shared prototypes was used by ViT-Net~\cite{kim2022vit}. Other studies focus on specific domains, such as the medical field, where the interpretability of models is especially important, \eg NP-ProtoPNet~\cite{singh2021these}, which operates on X-rays,
and ProtoBagNet~\cite{djoumessi2024actually}, which operates on Optical Coherence Tomography.

Besides the simple upscaling approach from ProtoPNet~\cite{chen2019looks}, there are more sophisticated methods to generate saliency maps, \eg Grad-CAM~\cite{selvaraju2017grad}, Integrated Gradients~\cite{sundararajan2017axiomatic}, LRP~\cite{montavon2019layer} or SmoothGrad~\cite{smilkov2017smoothgrad}. LRP was \eg adapted for the prototype context by Gautam et al.~\cite{gautam2023looks}, introducing Prototype-Relevance-Propagation (PRP). The PRP method is also used in our experiments to generate more truthful saliency maps.

With the emergence of increasingly powerful foundational models, the image segmentation domain has seen new opportunities~\cite{zhou2024image}. There has been extensive work to incorporate or extract information from foundational models such as CLIP~\cite{radford2021learning}, Stable Diffusion~\cite{rombach2022high}, and DINOv2~\cite{oquab2023dinov2}. Models directly focusing on image segmentation such as SAM~\cite{kirillov2023segment} and its successor SAM2~\cite{ravi2024sam2} are widely used as a basis for further research, such as the subobject-level image tokenization method SLIT from Chen et al.~\cite{chen2024subobject} or SEEM~\cite{zou2024segment}.  

General studies have demonstrated the complexity and diversity of quantative evaluation techniques employed in our context~\cite{nauta2023anecdotal}~\cite{zhou2021evaluating}. A more detailed survey specific to prototype models in the image domain is provided by Nauta et al.~\cite{nauta2023co}, with a practical demonstration of this evaluation framework in the study by Schlinge et al.~\cite{schlinge2025comprehensiveevaluationprototypeneural}. A major focus is a quantitative evaluation of the interpretability of prototypes in a semantic context, as assessed by Huang et al.~\cite{huang2023evaluation}.
As explainability is a second optimization goal, it is beneficial to incorporate this goal into the model tuning process, as demonstrated by the ProtoNeXt framework~\cite{willard2024looks}.

\begin{figure*}[!thb]
    \centering
    \includegraphics[width=.99\textwidth]{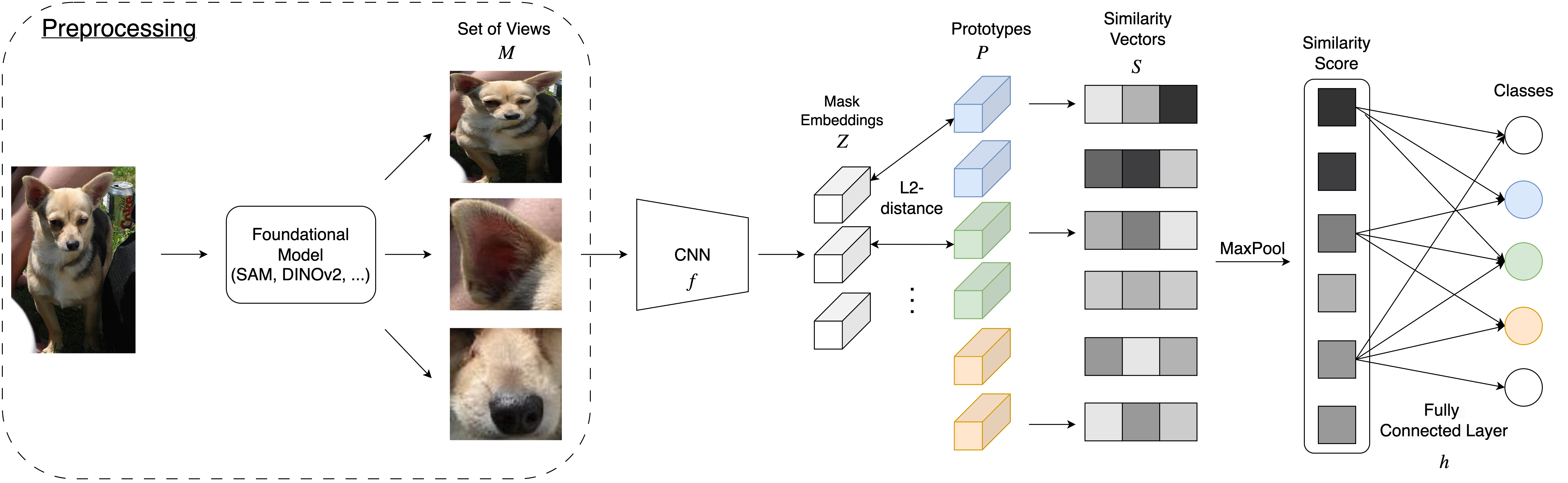}
    \caption{ProtoMask operates on multiple views of the same image, as illustrated on the left side. The design consists of a CNN backbone model, which extracts a feature vector $z$ for each view separately. A similarity score for each prototype feature vector pair is computed. The score of the nearest feature vector per prototype is then used in the classification layer $h$.}
    \label{fig:protomask_architecture}
\end{figure*}

\section{ProtoMask}
Our proposed model ProtoMask\footnote{Available at: \url{https://github.com/uos-sis/quanproto}} is based on ProtoPNet~\cite{chen2019looks} and adopts the design of class-distinct prototypes in a linear classification process.
Given an input image $x \in X$ from our dataset $(X,Y)$, we first compute a set of views $M_x = \{ m_i \in M_x : m_i \in \mathbb{R}^{3 \times H \times W}, i = 1,...,|M_x| \}$ using a foundational segmentation model, as shown in Figure~\ref{fig:protomask_architecture}. Each view is then passed through a feature extractor $f$, which is a CNN model, creating a set of embedding vectors $ Z_x = \{ z_i \in f(M_x) : z_i \in \mathbb{R}^D, i = 1,...,|M_x|\}$. By passing each view individually through the CNN model, we mitigate the challenges of large receptive fields to achieve a more truthful matching between embedding vector and input space. Subsequently, we calculate a similarity vector for each prototype $s_p = \{ \log(1 + \frac{1}{|| z_i - p||^{2}_{2}}) : z_i \in Z_x\}$ between the feature vectors $Z_x$ and the set of prototypes $P = \{ p_i \in P : p_i \in \mathbb{R}^D, i = 1, ..., |P| \}$ based on the L2-distance. The final step is to use the maximum similarity value of each similarity vector $\tilde{s}_p = \max(s_{p}) \in S_x$ as input in the linear classification process, represented by the fully connected layer $h$. The final output of the model is a probability distribution $o_c$ over the classes.

We combine different loss terms, targeting performance and explainability:
Starting with a cross entropy loss $\mathcal{L}_C$ for classification accuracy, we add the cluster loss $\mathcal{L}_{clst}$ and separation loss $\mathcal{L}_{sep}$, introduced by Chen et al.~\cite{chen2019looks} to promote a separated latent space, which aims to increase the explainability of the classification process. The final loss term is a diversity loss $\mathcal{L}_{div}$ on prototypes from the same class. We use the L2 distance to be consistent with the other loss terms.
\begin{align}
    \mathcal{L}_{clst}&=\frac{1}{n} \sum_{i=1}^n \min _{j: \mathbf{p}_j \in \mathbf{P}_{y_i}} \min_{\mathbf{z} \in \mathbf{Z}_x} \left\|\mathbf{z}-\mathbf{p}_j\right\|_2^2 \\
    \mathcal{L}_{sep}&=-\frac{1}{n} \sum_{i=1}^n \min _{j: \mathbf{p}_j \notin \mathbf{P}_{y_i}} \min _{\mathbf{z} \in \mathbf{Z}_x}\left\|\mathbf{z}-\mathbf{p}_j\right\|_2^2 \\
    \mathcal{L}_{div}&=\exp( - \alpha \sum_{i: \mathbf{p}_i \in \mathbf{P}_k}\sum_{j: \mathbf{p}_j \in \mathbf{P}_k \setminus \mathbf{p}_i} ||\mathbf{p}_i - \mathbf{p}_j||^{2}_{2})
\end{align} 
Related approaches to our diversity loss can be found in TesNet~\cite{wang2021interpretable}, ProtoPool~\cite{rymarczyk2022interpretable} or PIPNet~\cite{nauta2023pip} in different forms and are used to prevent the model from learning the same object property multiple times. Our overall learning objective $\mathcal{L}$ is given as $\mathcal{L}=\lambda_C \mathcal{L}_C+\lambda_{cl} \mathcal{L}_{clst}+\lambda_{sep} \mathcal{L}_{sep} + \lambda_{div} \mathcal{L}_{div}$, with hyperparameters $\lambda_C, \lambda_{cl}, \lambda_{sep}, \lambda_{div}$.

We employ the same two-stage training process as ProtoPNet~\cite{chen2019looks}. In the first stage, we train the feature extractor $f$ and the prototypes $P$, while fixing the fully connected layer $h$. The fixed fully connected layer is used to assign prototypes to classes, initialized with a value of 1 for class assignment and 0 otherwise. This initialization encourages the learning of class distinct prototypes via the classification loss. After the first stage, we project the prototypes to their nearest latent feature vector from one of the views in our training data.
\begin{equation}
p \leftarrow \underset{\mathbf{z} \in \mathbf{Z}_K}{\arg \min }\|\mathbf{z}-\mathbf{p}\|^{2}_{2}\,,
\end{equation}
where $Z_K=\left\{z_i: z_i \in Z_x \text { for all }(x, y): y \in K\right\}$. In the final stage, we finetune the fully connected layer $h$ on the classification task. Only the classification loss is used with an additional L1 regularization term to promote sparsity in the weight matrix of the fully connected layer. The backbone network $f$ and prototypes $P$ remain frozen during this stage.

\section{Evaluation Metrics}
The chosen models are evaluated with a subset of metrics from the QuanProto library~\cite{schlinge2025comprehensiveevaluationprototypeneural} to assess different aspects of explainability. The classification performance, is evaluated through the top-1 accuracy, top-3 accuracy and the F1 score.
%
Different compactness properties are assessed by comparing the global number of prototypes that are active in the model, defined as the \textit{Global Size}; the \textit{Sparsity} of the classification layer; and the negative-to-positive weight ratio of the classification layer, noted as \textit{NPR}.
%
To measure the quality of typical user explanations, we evaluate the top 5 most activated prototypes per test image. The first metric is the location change between prototype visualizations, defined as the \textit{VLC} score. The saliency method used to generate the visualizations is the Prototype-Relevance-Propagation (PRP) method, introduced by Gautam et al.~\cite{gautam2023looks}. Additionally, we assess the quality of the learned embedding space by quantifying the average cosine distance between prototypes of the same class and prototypes of different classes, using the $APD_{intra}$ and $APD_{inter}$ scores.
%
Another important explainability aspect is the covariant complexity of individual prototypes. The goal of the following metrics is to quantify the alignment between user expectations and the actual prototype properties. \textit{Object Overlap} and \textit{Background Overlap} measurements from visualizations are used to assess the alignment between the object and a prototype, which is further assessed with the inside-outside relevance distance, \textit{IORD}, evaluating the relevance scores computed on Object and Background. Finally, we assess the alignment of a prototype to a specific object part with the \textit{Consistency} measure. All metrics are defined and described in detail in~\cite{schlinge2025comprehensiveevaluationprototypeneural}.

\section{Experiments}
We selected CUB200-2011~\cite{wah2011caltech}, Stanford Dogs~\cite{khosla2011novel} and Stanford Cars~\cite{krause20133d} as three popular fine-grained classification datasets. This collection contains both natural and manufactured objects, which we expect to have different segmentation characteristics with our chosen methods. All datasets provide ground truth bounding boxes, and the CUB200 dataset additionally includes object masks and object-part annotations used for the covariant complexity metrics.

We choose to compare our model with ProtoPNet~\cite{chen2019looks}, ProtoPool~\cite{rymarczyk2022interpretable} and PIP-Net~\cite{nauta2023pip}.
We note that the described training procedure is not optimized for each model, and therefore a decline in performance is anticipated. Given the computational cost of fully tuning some models for new datasets, we prioritize a fair comparison under non-optimal settings.
\begin{figure}[!t]
    \begin{minipage}{\linewidth}
    \resizebox{\linewidth}{!}{
        \includegraphics[width=\linewidth]{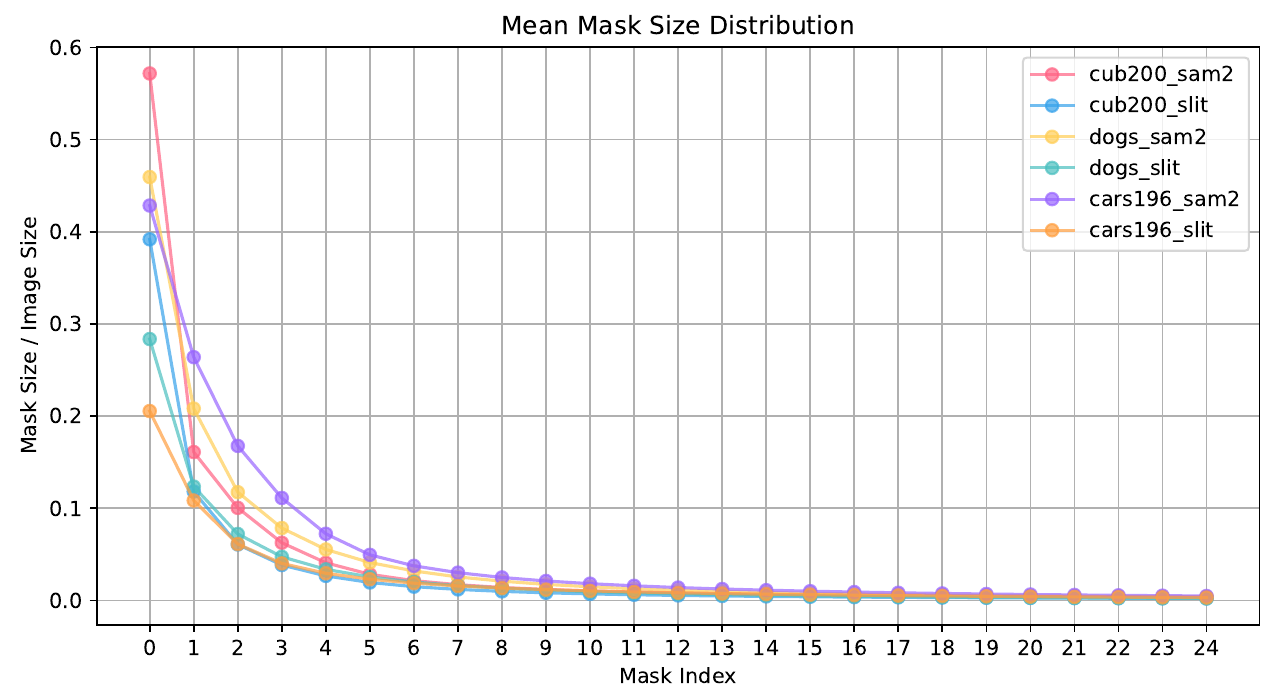}
    }
    \end{minipage}
    \begin{minipage}{\linewidth}
    \resizebox{\linewidth}{!}{
        \includegraphics[width=\linewidth]{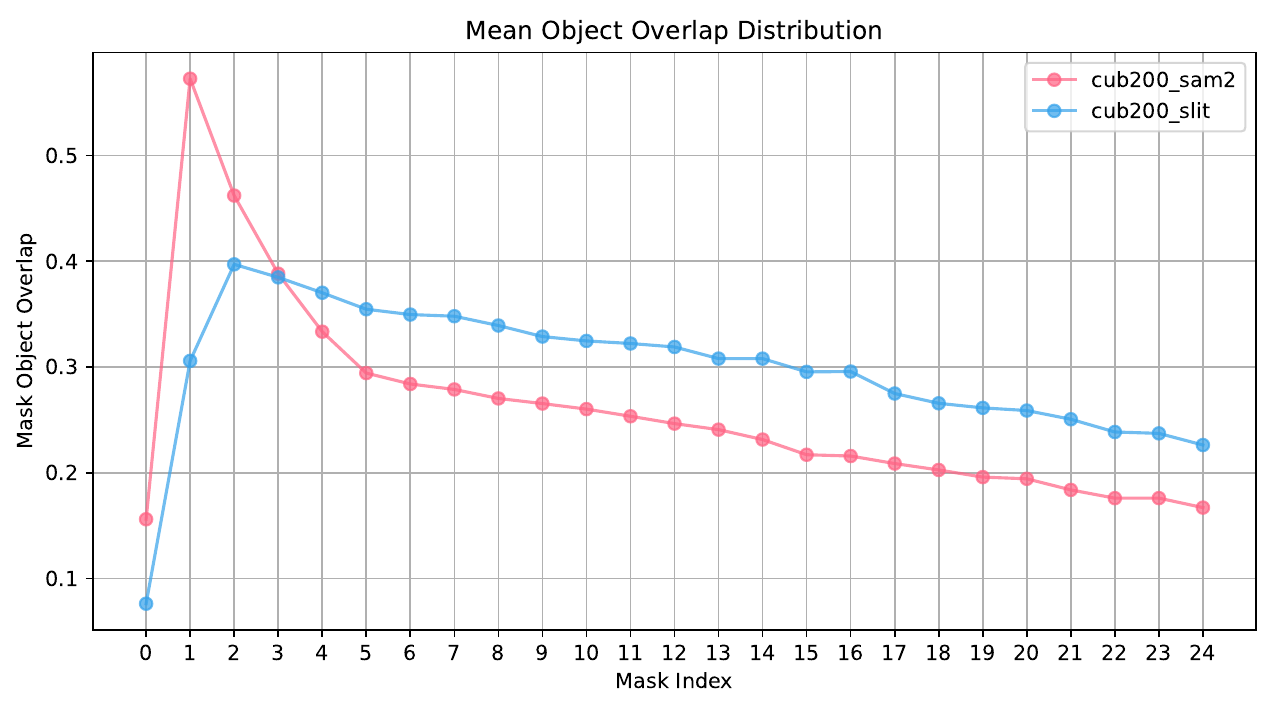}
    }
    \end{minipage}
    \caption{Distribution of the mean segmentation mask size (top) and mean mask object overlap (bottom) per image across all datasets for the SAM2 and SLIT segmentation method.}
    \label{fig:mask_distribution}
\end{figure}
All models, including ours, use a ResNet-50 backbone with pretrained feature weights on the ImageNet database. The decision to employ the ResNet architecture was based on the availability of a Prototype-Relevance-Propagation implementation for this model family. The feature map dimension for ProtoPNet and ProtoPool is set to $7 \times 7 \times 128$ across all datasets. PIPNet's CNN output tensor dimension is set to $7 \times 7 \times 2048$. PIPNet bypasses the similarity computation step used by other models and directly interprets this tensor as a collection of $7 \times 7$ similarity maps for $2048$ prototypes. The number of 2048 prototypes is used for all datasets. This aligns with PIPNet's design goal of learning a minimal prototype subset, thereby reducing the number of active prototypes measured with the \textit{Global Size} metric. Reducing this metric is not exclusive to the PIPNet model, as it is a general measure of the number of prototypes used in the classification process. ProtoPNet and our ProtoMask model use 10 prototypes per class, resulting in 2000, 1200, and 1960 prototypes for the CUB200, Dogs, and Cars datasets, respectively. For ProtoPool we choose a number slightly higher than the class count, consistent with the original work, resulting in 205, 125, and 201 prototypes for the CUB200, Dogs and Cars datasets, respectively.

We employed the default training setup from the QuanProto library~\cite{schlinge2025comprehensiveevaluationprototypeneural} with an online augmentation approach. A typical training routine can consist of 3 phases: a warm-up step adjusting prototype weights to the pretrained CNN weights, a joint step constituting the primary training phase, and a fine-tune step refining the final classification layer. All models are trained for 100 joint epochs with model-specific warm-up and fine-tune periods. The ProtoPNet, ProtoPool and PIPNet parameters for the CUB200 and Cars datasets are the defaults from the library. The parameters for the Dogs dataset were determined through hyperparameter tuning, with 50 trials for all models, optimizing accuracy.

Image preprocessing of ProtoPNet, ProtoPool and PIPNet includes the use of the ground truth bounding box information to crop the images to the object of interest. As our objective is to investigate the effect of cropping the image into semantically meaningful patches, we also trained the models without this preprocessing step, allowing for evaluation between a no zoom, object zoom, and object-part zoom setting. Model parameters for the non-cropped setting were also chosen based on hyperparameter tuning with 50 trails. Our segmentation process operates on the original image without ground truth bounding box cropping, for a realistic application scenario.

We selected the SAM2~\cite{ravi2024sam2} segmentation model as our first method, specifically the SAM2 B+ model. A key characteristic is its hierarchical mask structure, where a single image region can be included in multiple masks. The second segmentation model is SLIT~\cite{zou2024segment}, which does not directly output segmentation masks but predicts contour lines as a binary image. To segment the image, we resize these line predictions to the original image size and apply a dilation kernel of size 3 to close small gaps. A connected components' algorithm is then used to segment the image into individual masks. Note that the masks do not overlap, unlike SAM2.
%
The parameters for both methods were chosen based on hyperparameter tuning with 25 trials. We optimized for consistency of masks between images of the same class using a small subset of the CUB200 dataset and the object-part annotations. The mean occurrence of semantically identical masks in a class is $8.95\%$ and $7.48\%$ for SAM2 and SLIT, respectively. The possible influence of non-consistent mask sets is a major focus of our discussion, as we see a discrepancy with learning consistent prototypes. SAM2 parameters are \textit{points per side}: 32, \textit{crop n layer}: 2, and \textit{points downscale factor}: 1; other parameters are kept at their default values. The SLIT \textit{threshold} is set to 0.001. These parameters were applied to all datasets, regardless of differences between natural and manufactured objects.

To determine the number of views used for training for each method and dataset combination, we evaluated the decline in mask size within the set of masks generated from an image. We sorted each mask set by its percental image size and averaged the results over the entire dataset. The top diagram in Figure~\ref{fig:mask_distribution} shows this distribution of mean mask size for each dataset. The number of masks for a method-dataset pair is defined as the number of masks that reach an average size above 1\% of the image. The CUB200 dataset allowed us to further evaluate the average object overlap decline in the mask set, as shown in the bottom diagram of Figure~\ref{fig:mask_distribution}. The object overlap distribution demonstrates that small masks still retain object information; therefore, the limiting factor is the resolution problem arising from upscaling these small masks to generate our views. The resulting number of masks for SAM2 segmentations are 11,13 and 16 for the CUB200, Dogs and Cars datasets, respectively. For SLIT segmentations, we obtained 8, 10, and 11 for the CUB200, Dogs and Cars datasets, respectively.

\section{Results}
In this section, we present the results of our experiments. General performance, compactness and contrastivity metrics are evaluated on all datasets. The complexity evaluation is only conducted on the CUB200 dataset, because of missing object mask and part annotation ground truth information. The contrastivity and complexity evaluations use only the top-5 activated prototypes per image, which we see as a reasonable number for a comprehensive visual explanation to the user.
\begin{table*}[!t]
\centering
    \resizebox{0.8\linewidth}{!}{
        \begin{tabular}{lccccccccc}
            \toprule
            CUB200 & Accuracy$\%\uparrow$ & top-3 Acc.$\%\uparrow$  & F1 score$\%\uparrow$ & Global Size $\downarrow$ & Sparsity$\%\uparrow$ & NPR $\downarrow$ \\
            \midrule
            P.Mask SAM2 & 73.03 $\pm$ 0.51 & 86.02 $\pm$ 0.51 & 73.61 $\pm$ 0.45 & 2000 $\pm$ 0 & 99.47 $\pm$ 0.03 & 0.01 $\pm$ 0.02 \\
            P.Mask SLIT & 60.08 $\pm$ 0.93 & 77.35 $\pm$ 0.29 & 60.68 $\pm$ 1.18 & 2000 $\pm$ 0 & 99.44 $\pm$ 0.04 & 0.02 $\pm$ 0.02 \\
            PIPNet & \textbf{74.38} $\pm$ 0.33 & \textbf{87.40} $\pm$ 0.24 & \textbf{74.34} $\pm$ 0.35 & 1029 $\pm$ 73 & 99.06 $\pm$ 0.11 & \textbf{0.00} $\pm$ 0.00 \\
            PIPNet NC & 71.77 $\pm$ 0.18 & 85.00 $\pm$ 0.39 & 71.71 $\pm$ 0.24 & 575 $\pm$ 56 & \textbf{99.73} $\pm$ 0.05 & \textbf{0.00} $\pm$ 0.00 \\
            P.PNet & 68.62 $\pm$ 0.72 & 84.99 $\pm$ 0.48 & 68.85 $\pm$ 0.62 & 2000 $\pm$ 0 & 99.36 $\pm$ 0.03 & 0.16 $\pm$ 0.05 \\
            P.PNet NC & 63.47 $\pm$ 0.26 & 81.26 $\pm$ 0.50 & 63.94 $\pm$ 0.43 & 2000 $\pm$ 0 & 83.77 $\pm$ 1.10 & 0.82 $\pm$ 0.05 \\
            P.Pool & 67.74 $\pm$ 1.20 & 81.07 $\pm$ 0.68 & 68.22 $\pm$ 1.12 & \textbf{205} $\pm$ 0 & 97.41 $\pm$ 0.71 & 0.32 $\pm$ 0.22 \\
            P.Pool NC & 67.20 $\pm$ 0.37 & 77.67 $\pm$ 0.46 & 67.59 $\pm$ 0.40 & \textbf{205} $\pm$ 0 & 99.45 $\pm$ 0.04 & 0.01 $\pm$ 0.01 \\
            \bottomrule
            Dogs & Accuracy$\%\uparrow$ & top-3 Acc.$\%\uparrow$ & F1 score$\%\uparrow$ & Global Size $\downarrow$ & Sparsity$\%\uparrow$ & NPR $\downarrow$ \\
            \midrule
            P.Mask SAM2 & \textbf{79.38} $\pm$ 0.34 & \textbf{94.01} $\pm$ 0.22 & \textbf{79.41} $\pm$ 0.36 & 1200 $\pm$ 0 & 97.66 $\pm$ 0.75 & 0.15 $\pm$ 0.06 \\
            P.Mask SLIT & 67.68 $\pm$ 0.40 & 83.91 $\pm$ 0.58 & 68.13 $\pm$ 0.42 & 1200 $\pm$ 0 & 99.14 $\pm$ 0.02 & \textbf{0.00} $\pm$ 0.00 \\
            PIPNet & 77.98 $\pm$ 0.44 & 92.71 $\pm$ 0.87 & 77.87 $\pm$ 0.47 & 1309 $\pm$ 479 & 98.10 $\pm$ 1.03 & \textbf{0.00} $\pm$ 0.00 \\
            PIPNet NC & 73.66 $\pm$ 0.17 & 89.98 $\pm$ 0.72 & 73.57 $\pm$ 0.17 & 678 $\pm$ 73 & 98.74 $\pm$ 0.29 & \textbf{0.00} $\pm$ 0.00 \\
            P.PNet & 75.20 $\pm$ 2.49 & 91.25 $\pm$ 0.98 & 75.17 $\pm$ 2.42 & 1200 $\pm$ 0 & 97.11 $\pm$ 2.18 & 0.35 $\pm$ 0.32  \\
            P.PNet NC & 72.08 $\pm$ 0.17 & 89.25 $\pm$ 0.29 & 72.17 $\pm$ 0.11 & 1200 $\pm$ 0 & 91.70 $\pm$ 1.98 & 0.65 $\pm$ 0.10 \\
            P.Pool & 76.79 $\pm$ 0.79 & 87.81 $\pm$ 0.53 & 76.87 $\pm$ 0.77 & \textbf{125} $\pm$ 0 & 97.23 $\pm$ 0.50 & 0.32 $\pm$ 0.18 \\
            P.Pool NC & 72.05 $\pm$ 0.37 & 83.81 $\pm$ 0.56 & 72.06 $\pm$ 0.43 & \textbf{125} $\pm$ 0 & \textbf{99.16} $\pm$ 0.01 & \textbf{0.00} $\pm$ 0.00 \\
            \bottomrule
            Cars & Accuracy$\%\uparrow$ & top-3 Acc.$\%\uparrow$ & F1 score$\%\uparrow$ & Global Size $\downarrow$ & Sparsity$\%\uparrow$ & NPR $\downarrow$ \\
            \midrule
            P.Mask SAM2 & 80.79 $\pm$ 1.99 & 91.42 $\pm$ 0.91 & 80.96 $\pm$ 2.16 & 1960 $\pm$ 0 & 99.38 $\pm$ 0.11 & 0.05 $\pm$ 0.03 \\
            P.Mask SLIT & 64.65 $\pm$ 2.02 & 80.10 $\pm$ 0.99 & 65.15 $\pm$ 2.13 & 1960 $\pm$ 0 & 99.38 $\pm$ 0.09 & 0.03 $\pm$ 0.04 \\
            PIPNet & \textbf{86.24} $\pm$ 0.27 & \textbf{94.83} $\pm$ 0.40 & \textbf{86.17} $\pm$ 0.23 & 496 $\pm$ 22 & \textbf{99.50} $\pm$ 0.03 & \textbf{0.00} $\pm$ 0.00 \\
            PIPNet NC & 82.24 $\pm$ 0.49 & 92.49 $\pm$ 0.48 & 82.14 $\pm$ 0.54 & 720 $\pm$ 67 & 99.19 $\pm$ 0.22 & \textbf{0.00} $\pm$ 0.00 \\
            P.PNet & 81.89 $\pm$ 1.23 & 93.90 $\pm$ 0.65 & 81.87 $\pm$ 1.22 & 1960 $\pm$ 0 & 99.23 $\pm$ 0.09 & 0.14 $\pm$ 0.07 \\
            P.PNet NC & 73.37 $\pm$ 1.67 & 89.54 $\pm$ 0.56 & 73.63 $\pm$ 1.66 & 1960 $\pm$ 0 & 99.08 $\pm$ 0.31 & 0.42 $\pm$ 0.21 \\
            P.Pool & 81.31 $\pm$ 0.96 & 90.35 $\pm$ 0.87 & 81.32 $\pm$ 0.92 & \textbf{201} $\pm$ 0 & 99.12 $\pm$ 0.21 & 0.11 $\pm$ 0.08 \\
            P.Pool NC & 77.07 $\pm$ 0.98 & 84.94 $\pm$ 0.64 & 77.52 $\pm$ 0.91 & \textbf{201} $\pm$ 0 & 99.31 $\pm$ 0.10 & 0.13 $\pm$ 0.07 \\
            \bottomrule
            \end{tabular}
    }
    \caption{General performance and compactness results, obtained from 4 independent runs with different seeds to calculate the mean and standard deviation. ``NC'' indicates preprocessing without cropping.}
    \label{tab:general_compact}
\end{table*}

The results of our general performance and compactness evaluation are illustrated in Table \ref{tab:general_compact}. The abbreviation NC represents the learning setup with no cropping in the preprocessing stage. PIPNet demonstrates robust performance across all datasets. The performance of our ProtoMask model indicates quality differences in the generated segmentation masks between methods. Masks generated using the SLIT method generally exhibit worse performance. Furthermore, results indicate a notable difference in mask quality between natural and manufactured objects, as observed by comparing the performance ranking of CUB200 and Dogs to the Cars ranking. Except for ProtoPool NC, removing the cropping preprocessing step reduces the performance of models. This demonstrates the general impact of object size in the image on performance. This suggests that cropping to the object of interest generally increases the performance, which supports our study on more sophisticated cropping strategies. The ProtoPNet and ProtoPool models generally have lower performance with higher standard deviation on some datasets.

PIPNet is the only model capable of reducing the \textit{Global Size} across all datasets that is measured by the number of prototypes used for classification. Removing the cropping preprocessing step for PIPNet yields mixed results regarding \textit{Global Size}, with natural objects exhibiting lower and manufactured objects exhibiting higher \textit{Global Size} scores compared to when cropping is included. The other models, including ours, cannot reduce the \textit{Global Size}, indicating that the L1-regulation term used in the fine tune phase does not affect the number of used prototypes in the classification layer in the current training and parameter setup.
The ProtoMask model achieves robust \textit{Sparsity} scores comparable to the top scores of each dataset. All models achieved a \textit{Sparsity} above 99\% on the Cars dataset; however, the CUB200 dataset shows two exceptions to this, and on the Dogs dataset, the majority of models uses larger sets of prototypes relative to the number of classes. Due to PIPNet's hard design constraint preventing negative weights in the classification layer, the model always achieves a perfect \textit{NPR} score. ProtoPNet and ProtoPool struggle to prevent negative weights in the classification layer. The ProtoMask modification of the ProtoPNet design facilitates the prevention of negative weights, increasing the explainability of the classification process.

\begin{table*}[!t]
    \centering
    \resizebox{\linewidth}{!}{
        \begin{tabular}{lcccccccc}
            \toprule
            CUB200 & $VLC \uparrow$ & $APD_{intra} \uparrow$  & $APD_{inter} \uparrow$ & Object Overlap $\%\uparrow$ & Background Overlap $\%\downarrow$ &  IORD $\uparrow$  & Consistency $\%\uparrow$  \\
            \midrule
            P.Mask SAM2 & 21.58 $\pm$ 3.33 & 0.00 $\pm$ 0.00 & \textbf{0.02} $\pm$ 0.00 & \textbf{42.61} $\pm$ 2.65 & 29.57 $\pm$ 7.44 & 0.03 $\pm$ 0.01 & 6.08 $\pm$ 0.92 \\
            P.Mask SLIT & 16.87 $\pm$ 0.45 & 0.00 $\pm$ 0.00 & 0.01 $\pm$ 0.00 & 42.55 $\pm$ 0.22 & 29.48 $\pm$ 0.59 & \textbf{0.05} $\pm$ 0.00 & 6.02 $\pm$ 0.38 \\
            PIPNet & \textbf{86.81} $\pm$ 0.20 & - & - & 5.01 $\pm$ 0.22 & 62.19 $\pm$ 1.60 & -0.02 $\pm$ 0.00 & 48.32 $\pm$ 1.45 \\
            PIPNet NC & 83.53 $\pm$ 0.18 & - & - & 9.86 $\pm$ 0.43 & 77.20 $\pm$ 0.99 & -0.04 $\pm$ 0.00 & 45.55 $\pm$ 1.28 \\
            P.PNet & 54.98 $\pm$ 1.38 & 0.00 $\pm$ 0.00 & 0.01 $\pm$ 0.01 & 10.14 $\pm$ 2.01 & \textbf{25.83} $\pm$ 14.66 & 0.03 $\pm$ 0.03 & 47.08 $\pm$ 8.3 \\
            P.PNet NC & 52.95 $\pm$ 5.84 & 0.00 $\pm$ 0.00 & \textbf{0.02} $\pm$ 0.00 & 21.57 $\pm$ 3.83 & 55.65 $\pm$ 7.62 & 0.03 $\pm$ 0.02 & \textbf{63.94} $\pm$ 0.76 \\
            P.Pool & 28.45 $\pm$ 25.92 & 0.00 $\pm$ 0.00 & 0.00 $\pm$ 0.00 & 3.65 $\pm$ 3.71 & 72.80 $\pm$ 26.82 & -0.05 $\pm$ 0.04 & 31.26 $\pm$ 5.93 \\
            P.Pool NC & 50.63 $\pm$ 18.51 & \textbf{0.01} $\pm$ 0.00 & 0.01 $\pm$ 0.00 & 21.82 $\pm$ 11.80 & 54.57 $\pm$ 23.89 & 0.01 $\pm$ 0.05 & 54.5 $\pm$ 3.59 \\
            \end{tabular}
    }
    \resizebox{0.9\linewidth}{!}{
        \begin{tabular}{lccc|lccc}
            \toprule
            Dogs & $VLC \uparrow$ & $APD_{intra} \uparrow$  & $APD_{inter} \uparrow$ &  Cars & $VLC \uparrow$ & $APD_{intra} \uparrow$  & $APD_{inter} \uparrow$ \\
            \midrule
            P.Mask SAM2 & 20.30 $\pm$ 0.73 & 0.00 $\pm$ 0.00 & 0.00 $\pm$ 0.00 & P.Mask SAM2 &       20.22 $\pm$ 1.70 & 0.00 $\pm$ 0.00 & 0.01 $\pm$ 0.01 \\
            P.Mask SLIT & 25.16 $\pm$ 2.90 & 0.00 $\pm$ 0.00 & \textbf{0.02} $\pm$ 0.00 & P.Mask SLIT &       17.30 $\pm$ 1.16 & 0.00 $\pm$ 0.00 & 0.01 $\pm$ 0.00 \\
            PIPNet & 81.90 $\pm$ 0.68 & - & - & PIPNet &                                        \textbf{86.01} $\pm$ 0.37 & - & - \\
            PIPNet NC & \textbf{84.77} $\pm$ 0.23 & - & - & PIPNet NC &                                 84.35 $\pm$ 0.64 & - & - \\
            P.PNet & 54.22 $\pm$ 8.82 & 0.00 $\pm$ 0.00 & \textbf{0.02} $\pm$ 0.02 & P.PNet &            58.91 $\pm$ 3.50 & 0.00 $\pm$ 0.00 & \textbf{0.03} $\pm$ 0.00 \\
            P.PNet NC & 54.19 $\pm$ 1.60 & 0.00 $\pm$ 0.00 & 0.01 $\pm$ 0.00 & P.PNet NC &     52.75 $\pm$ 1.86 & 0.00 $\pm$ 0.00 & 0.02 $\pm$ 0.00 \\
            P.Pool & 62.59 $\pm$ 14.48 & \textbf{0.01} $\pm$ 0.01 & 0.01 $\pm$ 0.01 & P.Pool &           52.98 $\pm$ 24.07 & \textbf{0.01} $\pm$ 0.00 & 0.01 $\pm$ 0.01 \\
            P.Pool NC & 47.70 $\pm$ 2.85 & \textbf{0.01} $\pm$ 0.00 & 0.01 $\pm$ 0.00 & P.Pool NC &     11.04 $\pm$ 3.02 & 0.00 $\pm$ 0.00 & 0.00 $\pm$ 0.00 \\
            \bottomrule
            \end{tabular}
    }
    \caption{Contrastivity and complexity results, averaged over 4 runs with different seeds. Absence of data indicates that the model architecture does not include the specific component being evaluated. ``NC'' indicates preprocessing without cropping.}
    \label{tab:contrastivity}
\end{table*}
Our contrastivity and complexity results are shown in Table \ref{tab:contrastivity}. PIPNet achieves the highest \textit{VLC} score over all datasets, demonstrating a robust prototype location contrast in the input space. This confirms that the design and optimization of the model actively improves this property. ProtoMask exhibits the lowest location change between prototypes of all models, with no clear ranking between the SAM2 and SLIT methods. This indicates that the model focuses only on a small subset of views in the training process, possibly due to low masks consistencies over the sample images. Other models fall between these two, with some dataset-dependent outliers. Indicating that the design and training process of the ProtoPNet and ProtoPool models only facilitate location contrast between prototypes without robust optimization. No embedding space clustering could be measured with the $APD_{intra}$ and $APD_{inter}$ scores, suggesting a strong dependency on learning parameters and training procedure, despite active optimization with dedicated loss functions.

ProtoMask achieves the highest \textit{Object Overlap}, indicating that the visual explanations of our prototypes cover large parts of the object. The \textit{Background Overlap} is one of the smallest, which means the model focuses on views picturing the object and ignores most views with significant background information. Due to our cropped views, this also suggests a lower risk of undesired phenomena like the clever-Hans problem. The only other model with a lower background coverage is the ProtoPNet model, but it shows high standard deviations. The other models achieve a detailed focus on object parts but focus mostly on the background, indicating a strong focus on background information for the classification process. The negative \textit{IORD} score supports the argument that PIPNet and ProtoPool focus mostly on the background. However, all \textit{IORD} are around zero, which means overall relevance amplitudes on object and background are similar. A possible scenario of a near zero \textit{IORD} score is when a prototype focuses mainly on the outline of the object. ProtoMask exhibits the lowest scores of object part \textit{Consistency} with both segmentation methods. This underscores the issue of inconsistent segmentations between images of the same class, as our results are lower than the average mask consistency of the SAM2 and SLIT method. Other models show notable higher \textit{Consistency} scores. However, these results also indicate that the learned prototypes represent multiple object-part or characteristics.

\section{Discussion}
\begin{figure}[!b]
\centering
    \resizebox{1.0\linewidth}{!}{
    \centering
    \begin{tabular}{cc} 
    \includegraphics[width=0.49\linewidth]{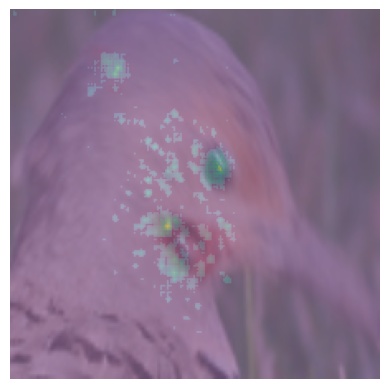} &
    \includegraphics[width=0.49\linewidth]{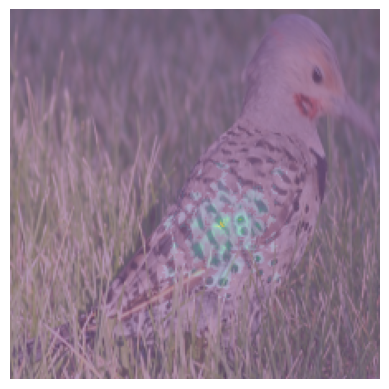} \\
    ProtoMask & PIPNet \\ 
    \includegraphics[width=0.49\linewidth]{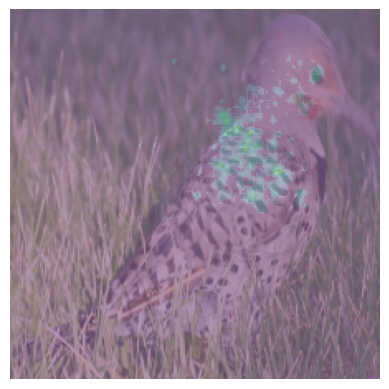} &
    \includegraphics[width=0.49\linewidth]{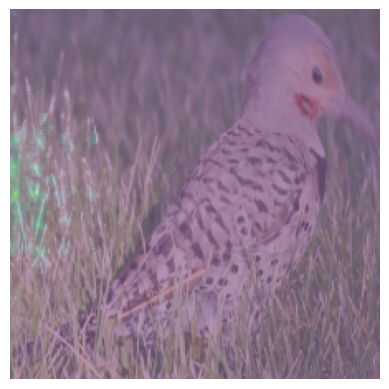} \\
    ProtoPNet & ProtoPool \\
    \end{tabular}
    }
    \caption{Prototype Relevance Propagation (PRP) visualization of one of the top-5 most activated prototypes on a representative test image across all models.}
    \label{fig:visualizations:top5}
\end{figure}

In the comparison of the prototype visualizations in Figure~\ref{fig:visualizations:top5}, the main benefits of our method regarding the interpretation of explanations is evident. The prototype visualization of the PIPNet and ProtoPool model can be associated with a single attribute, for example, the pattern of the feathers or the grass in the background. However, the ProtoPNet visualization illustrates the challenges we face with saliency methods. The visualization is more dispersed over the image, not focusing on a clear attribute. The cause of this dispersion remains unclear; it is uncertain whether the saliency method is imprecise or if this accurately reflects what the prototype has learned. Our approach reduces this uncertainty because the visualization is pre-focused on an object part, the head, mitigating the impact of potential inaccuracies of the saliency method. 

The results indicate a performance decline when using a different training routine than the original work. Some major changes in the PIPNet model were the decision to use a $7 \times 7$ spatial dimension for the CNN output, whereas the original work used $28 \times 28$. In addition, we use a prototype dimension of $1 \times 1 \times 128$ for ProtoPNet and ProtoPool across all datasets, for robust and fast training, as other dimensions introduced instability in our training setup. Nonetheless, as we focused on tuning and training the models under equal conditions, we argue that the results can be used for a representative comparison. The results demonstrate a performance decline without using the cropping preprocessing step on the other models. This was also stated in the original ProtoPNet work~\cite{chen2019looks}. Techniques like increasing the resolution of the CNN output and cropping the object beforehand, focus prototypes on smaller image regions. This is in line with our approach of modelling a more sophisticated cropping strategy to increase overall performance and decrease interpretation uncertainties in visual explanations.

\begin{figure*}[!t]
    \centering
    \resizebox{1.0\linewidth}{!}{
    \begin{tabular}{cc|c|c|c|c|c}

    &\multicolumn{2}{c|}{CUB200} & \multicolumn{2}{c|}{Stanford Dogs} &\multicolumn{2}{c}{Stanford Cars} \\
    \midrule
    a) & \raisebox{-.5\height}{\includegraphics[width=0.2\linewidth]{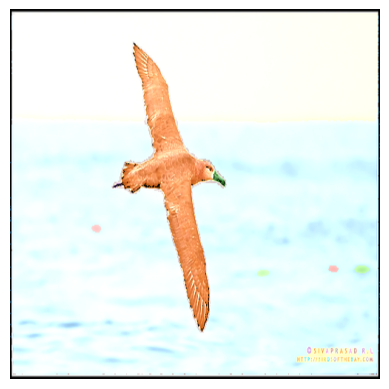}} &
    \raisebox{-.5\height}{\includegraphics[width=0.2\linewidth]{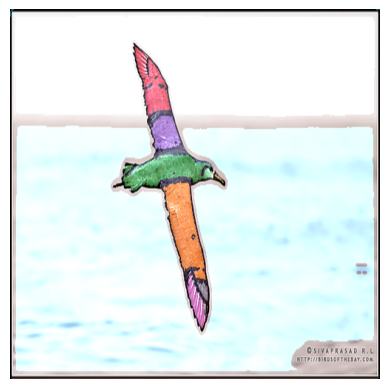}} &
    \raisebox{-.5\height}{\includegraphics[width=0.2\linewidth]{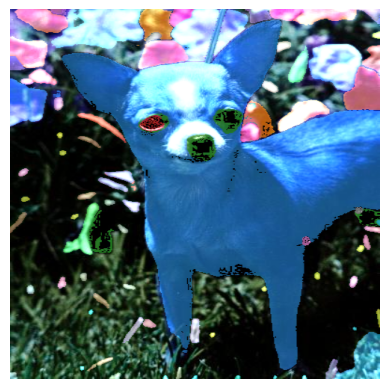}} &
    \raisebox{-.5\height}{\includegraphics[width=0.2\linewidth]{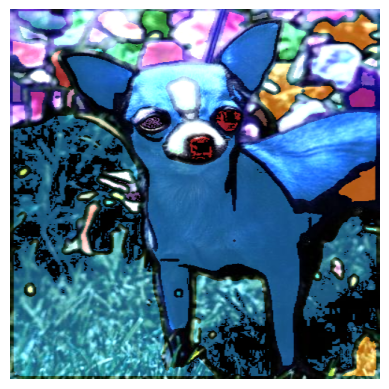}} &
    \raisebox{-.5\height}{\includegraphics[width=0.2\linewidth]{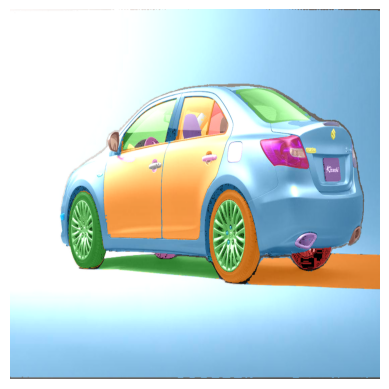}} &
    \raisebox{-.5\height}{\includegraphics[width=0.2\linewidth]{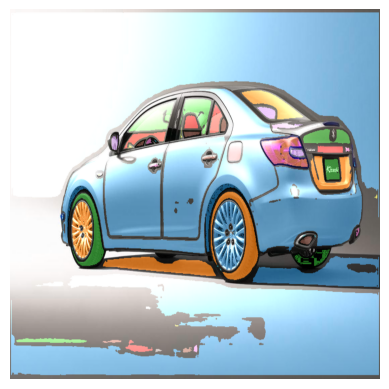}} \\
    \midrule
     b)&\raisebox{-.5\height}{\includegraphics[width=0.2\linewidth]{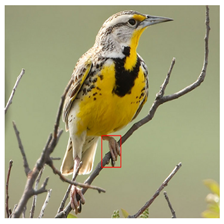}} &
    \raisebox{-.5\height}{\includegraphics[width=0.2\linewidth]{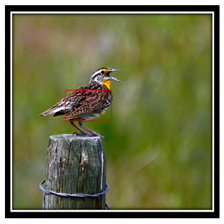}} &
    \raisebox{-.5\height}{\includegraphics[width=0.2\linewidth]{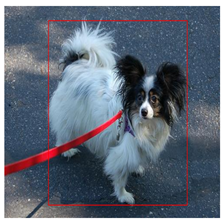}} &
    \raisebox{-.5\height}{\includegraphics[width=0.2\linewidth]{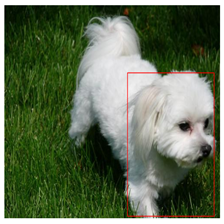}} &
    \raisebox{-.5\height}{\includegraphics[width=0.2\linewidth]{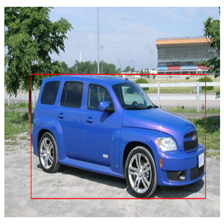}} &
    \raisebox{-.5\height}{\includegraphics[width=0.2\linewidth]{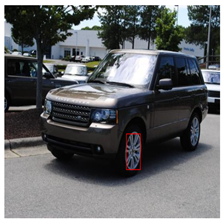}} \\
    \midrule
    &SAM2 & SLIT &SAM2 &SLIT &SAM2 &SLIT\\
    \end{tabular}
    }
    \caption{a) SAM2 and SLIT example segmentations across datasets. b) Examples of uncropped input views. 
    The illustrated examples were used in the projection step of our model, establishing a 1:1 relation between a prototype and its nearest input view.}
    \label{fig:visualizations}
\end{figure*}

A clear difference in ranking between natural and manufactured objects was observed for our model. Another notable result is the poor performance on segmentation masks generated with SLIT. This suggests that the used segmentation method needs to achieve specific properties. Figure~\ref{fig:visualizations} shows example segmentations for the two methods across all datasets. An observation is that SLIT appears to produce finer segmentations. Consequently, we hypothesize that highly detailed but potentially less consistent segmentations decrease the performance of our model.

We confirm the results from Nauta et al.~\cite{nauta2023pip} showing a reduction in \textit{Global Size} and a high \textit{Sparsity} of around 99\% for the PIPNet model, and no reduction in \textit{Global Size} for other models. Focusing on the \textit{Sparsity} and \textit{NPR} metrics that measure the complexity of the classification process, it can be seen that our adaption increases the explainability of the original ProtoPNet approach and shows overall one of the best results in this property, besides the PIPNet model with hard constrains.

Our model's low prototype location contrast in the input image supports our claim on quality difficulties of segmentation masks. The results can be interpreted as an indication that only a small number of consistent masks, for example, the entire object, could be found by the segmentation method. This results in poor location contrast due to a large number of duplicate prototypes. A distance between prototypes of the same class and different classes could not be confirmed in our $APD_{intra}$ and $APD_{inter}$ measurements. We argue that this property is highly influenced by the training process, for example, different learning rates in warm-up and joint phase, as we used the same loss weights from the original works.

The complexity results confirm the stated difficulties. The used segmentation masks include no consistent small-scale object parts. Instead, large segmentation masks covering a significant portion of the object are represented by the prototypes. Examples for this claim are provided in Figure~\ref{fig:visualizations}, in which uncropped input views used in the projection for individual prototypes are shown. It can also be argued that because of this segmentation inconsistency, prototypes tend to represent multiple masks to increase the performance of the model. This results in low \textit{Consistency} scores, and hinders the model to optimize contrastivity properties. Despite these conditions, our model still focuses mainly on the object itself and neglects background information better than the compared models. 

\section{Conclusion}
This work presents a novel architecture named ProtoMask studying potential improvements in performance and explainability using a multi-view learning approach generated through segmentation methods. Our findings suggest that mask characteristics of current state-of-the-art segmentation algorithms like SAM2 and SLIT are not well suited for the application of semantic object segmentation. However, even non-optimal segmentations can lead to clear performance improvements in some metrics, as our results demonstrate. Identifying the improvement of consistent segmentations under different perspectives and scenarios is a promising avenue for future research, which can greatly benefit the domain of prototype-based models, particularly concerning the explainability challenges these models face with the use of saliency maps. Exploring the latter further is another interesting direction for future research.

\bibliographystyle{IEEEtran}
\bibliography{references}

\begin{thebibliography}{10}
\providecommand{\url}[1]{#1}
\csname url@samestyle\endcsname
\providecommand{\newblock}{\relax}
\providecommand{\bibinfo}[2]{#2}
\providecommand{\BIBentrySTDinterwordspacing}{\spaceskip=0pt\relax}
\providecommand{\BIBentryALTinterwordstretchfactor}{4}
\providecommand{\BIBentryALTinterwordspacing}{\spaceskip=\fontdimen2\font plus
\BIBentryALTinterwordstretchfactor\fontdimen3\font minus
  \fontdimen4\font\relax}
\providecommand{\BIBforeignlanguage}[2]{{%
\expandafter\ifx\csname l@#1\endcsname\relax
\typeout{** WARNING: IEEEtran.bst: No hyphenation pattern has been}%
\typeout{** loaded for the language `#1'. Using the pattern for}%
\typeout{** the default language instead.}%
\else
\language=\csname l@#1\endcsname
\fi
#2}}
\providecommand{\BIBdecl}{\relax}
\BIBdecl

\bibitem{minh2022explainable}
D.~Minh, H.~X. Wang, Y.~F. Li, and T.~N. Nguyen, ``Explainable artificial
  intelligence: a comprehensive review,'' \emph{Artif. Intell. Rev.}, pp.
  1--66, 2022.

\bibitem{AFKS:24}
M.~Atzmueller, J.~Fürnkranz, T.~Kliegr, and U.~Schmid, ``{Explainable and
  Interpretable Machine Learning and Data Mining},'' \emph{Data Mining and
  Knowledge Discovery}, 2024.

\bibitem{chen2019looks}
C.~Chen, O.~Li, D.~Tao, A.~Barnett, C.~Rudin, and J.~K. Su, ``This looks like
  that: deep learning for interpretable image recognition,'' \emph{Advances in
  neural information processing systems}, vol.~32, 2019.

\bibitem{hoffmann2021looks}
A.~Hoffmann, C.~Fanconi, R.~Rade, and J.~Kohler, ``This looks like that... does
  it? shortcomings of latent space prototype interpretability in deep
  networks,'' \emph{arXiv preprint arXiv:2105.02968}, 2021.

\bibitem{selvaraju2017grad}
R.~R. Selvaraju, M.~Cogswell, A.~Das, R.~Vedantam, D.~Parikh, and D.~Batra,
  ``Grad-cam: Visual explanations from deep networks via gradient-based
  localization,'' in \emph{Proc. IEEE International Conference on Computer
  Vision}, 2017, pp. 618--626.

\bibitem{rymarczyk2022interpretable}
D.~Rymarczyk, {\L}.~Struski, M.~G{\'o}rszczak, K.~Lewandowska, J.~Tabor, and
  B.~Zieli{\'n}ski, ``Interpretable image classification with differentiable
  prototypes assignment,'' in \emph{Proc. European Conference on Computer
  Vision}.\hskip 1em plus 0.5em minus 0.4em\relax Springer, 2022, pp. 351--368.

\bibitem{nauta2023pip}
M.~Nauta, J.~Schl{\"o}tterer, M.~Van~Keulen, and C.~Seifert, ``Pip-net:
  Patch-based intuitive prototypes for interpretable image classification,'' in
  \emph{Proc. IEEE/CVF Conference on Computer Vision and Pattern Recognition},
  2023, pp. 2744--2753.

\bibitem{donnelly2022deformable}
J.~Donnelly, A.~J. Barnett, and C.~Chen, ``Deformable protopnet: An
  interpretable image classifier using deformable prototypes,'' in \emph{Proc.
  IEEE/CVF Conference on Computer Vision and Pattern Recognition}, 2022, pp.
  10\,265--10\,275.

\bibitem{gautam2023looks}
S.~Gautam, M.~M.-C. H{\"o}hne, S.~Hansen, R.~Jenssen, and M.~Kampffmeyer,
  ``This looks more like that: Enhancing self-explaining models by prototypical
  relevance propagation,'' \emph{Pattern Recognition}, vol. 136, p. 109172,
  2023.

\bibitem{xu2023sanity}
R.~Xu-Darme, G.~Qu{\'e}not, Z.~Chihani, and M.-C. Rousset, ``Sanity checks and
  improvements for patch visualisation in prototype-based image
  classification,'' \emph{arXiv preprint arXiv:2302.08508}, 2023.

\bibitem{neukirch2025featinv}
N.~Neukirch, J.~Vielhaben, and N.~Strodthoff, ``Featinv: Spatially resolved
  mapping from feature space to input space using conditional diffusion
  models,'' \emph{arXiv preprint arXiv:2505.21032}, 2025.

\bibitem{sun2024explain}
A.~Sun, P.~Ma, Y.~Yuan, and S.~Wang, ``Explain any concept: Segment anything
  meets concept-based explanation,'' \emph{Advances in Neural Information
  Processing Systems}, vol.~36, 2024.

\bibitem{kong2024lce}
W.~Kong, X.~Gong, and J.~Wang, ``Lce: A framework for explainability of dnns
  for ultrasound image based on concept discovery,'' \emph{arXiv preprint
  arXiv:2408.09899}, 2024.

\bibitem{kim2024vision}
Y.~K. Kim, J.~M. Di~Martino, and G.~Sapiro, ``Vision transformers with natural
  language semantics,'' \emph{arXiv preprint arXiv:2402.17863}, 2024.

\bibitem{wang2023learning}
C.~Wang, Y.~Liu, Y.~Chen, F.~Liu, Y.~Tian, D.~McCarthy, H.~Frazer, and
  G.~Carneiro, ``Learning support and trivial prototypes for interpretable
  image classification,'' in \emph{Proc. IEEE/CVF International Conference on
  Computer Vision}, 2023, pp. 2062--2072.

\bibitem{rymarczyk2021protopshare}
D.~Rymarczyk, {\L}.~Struski, J.~Tabor, and B.~Zieli{\'n}ski, ``Protopshare:
  Prototypical parts sharing for similarity discovery in interpretable image
  classification,'' in \emph{Proc. ACM SIGKDD}, 2021, pp. 1420--1430.

\bibitem{hase2019interpretable}
P.~Hase, C.~Chen, O.~Li, and C.~Rudin, ``Interpretable image recognition with
  hierarchical prototypes,'' in \emph{Proc. AAAI Conference on Human
  Computation and Crowdsourcing}, vol.~7, 2019, pp. 32--40.

\bibitem{nauta2021neural}
M.~Nauta, R.~Van~Bree, and C.~Seifert, ``Neural prototype trees for
  interpretable fine-grained image recognition,'' in \emph{Proc. IEEE/CVF
  Conference on Computer Vision and Pattern Recognition}, 2021, pp.
  14\,933--14\,943.

\bibitem{xue2022protopformer}
M.~Xue, Q.~Huang, H.~Zhang, L.~Cheng, J.~Song, M.~Wu, and M.~Song,
  ``Protopformer: Concentrating on prototypical parts in vision transformers
  for interpretable image recognition,'' \emph{arXiv preprint
  arXiv:2208.10431}, 2022.

\bibitem{ma2024interpretable}
C.~Ma, J.~Donnelly, W.~Liu, S.~Vosoughi, C.~Rudin, and C.~Chen, ``Interpretable
  image classification with adaptive prototype-based vision transformers,'' in
  \emph{Proc. Annual Conference on Neural Information Processing Systems},
  2024.

\bibitem{kim2022vit}
S.~Kim, J.~Nam, and B.~C. Ko, ``Vit-net: Interpretable vision transformers with
  neural tree decoder,'' in \emph{Proc. International Conference on Machine
  Learning}.\hskip 1em plus 0.5em minus 0.4em\relax PMLR, 2022, pp.
  11\,162--11\,172.

\bibitem{singh2021these}
G.~Singh and K.-C. Yow, ``These do not look like those: An interpretable deep
  learning model for image recognition,'' \emph{IEEE Access}, vol.~9, pp.
  41\,482--41\,493, 2021.

\bibitem{djoumessi2024actually}
K.~Djoumessi, B.~Bah, L.~K{\"u}hlewein, P.~Berens, and L.~Koch, ``This actually
  looks like that: Proto-bagnets for local and global
  interpretability-by-design,'' in \emph{Proc. International Conference on
  Medical Image Computing and Computer-Assisted Intervention}.\hskip 1em plus
  0.5em minus 0.4em\relax Springer, 2024, pp. 718--728.

\bibitem{sundararajan2017axiomatic}
M.~Sundararajan, A.~Taly, and Q.~Yan, ``Axiomatic attribution for deep
  networks,'' in \emph{International conference on machine learning}.\hskip 1em
  plus 0.5em minus 0.4em\relax PMLR, 2017, pp. 3319--3328.

\bibitem{montavon2019layer}
G.~Montavon, A.~Binder, S.~Lapuschkin, W.~Samek, and K.-R. M{\"u}ller,
  ``Layer-wise relevance propagation: an overview,'' \emph{Explainable AI:
  interpreting, explaining and visualizing deep learning}, pp. 193--209, 2019.

\bibitem{smilkov2017smoothgrad}
D.~Smilkov, N.~Thorat, B.~Kim, F.~Vi{\'e}gas, and M.~Wattenberg, ``Smoothgrad:
  removing noise by adding noise,'' \emph{arXiv preprint arXiv:1706.03825},
  2017.

\bibitem{zhou2024image}
T.~Zhou, F.~Zhang, B.~Chang, W.~Wang, Y.~Yuan, E.~Konukoglu, and D.~Cremers,
  ``Image segmentation in foundation model era: A survey,'' \emph{arXiv
  preprint arXiv:2408.12957}, 2024.

\bibitem{radford2021learning}
A.~Radford, J.~W. Kim, C.~Hallacy, A.~Ramesh, G.~Goh, S.~Agarwal, G.~Sastry,
  A.~Askell, P.~Mishkin, J.~Clark \emph{et~al.}, ``Learning transferable visual
  models from natural language supervision,'' in \emph{Proc. International
  Conference on Machine Learning}.\hskip 1em plus 0.5em minus 0.4em\relax PMLR,
  2021, pp. 8748--8763.

\bibitem{rombach2022high}
R.~Rombach, A.~Blattmann, D.~Lorenz, P.~Esser, and B.~Ommer, ``High-resolution
  image synthesis with latent diffusion models,'' in \emph{Proc. IEEE/CVF
  Conference on Computer Vision and Pattern Recognition}, 2022, pp.
  10\,684--10\,695.

\bibitem{oquab2023dinov2}
M.~Oquab, T.~Darcet, T.~Moutakanni, H.~Vo, M.~Szafraniec, V.~Khalidov,
  P.~Fernandez, D.~Haziza, F.~Massa, A.~El-Nouby \emph{et~al.}, ``Dinov2:
  Learning robust visual features without supervision,'' \emph{arXiv preprint
  arXiv:2304.07193}, 2023.

\bibitem{kirillov2023segment}
A.~Kirillov, E.~Mintun, N.~Ravi, H.~Mao, C.~Rolland, L.~Gustafson, T.~Xiao,
  S.~Whitehead, A.~C. Berg, W.-Y. Lo \emph{et~al.}, ``Segment anything,'' in
  \emph{Proc. IEEE/CVF International Conference on Computer Vision}, 2023, pp.
  4015--4026.

\bibitem{ravi2024sam2}
N.~Ravi, V.~Gabeur, Y.-T. Hu, R.~Hu, C.~Ryali, T.~Ma, H.~Khedr, R.~R{\"a}dle,
  C.~Rolland, L.~Gustafson, E.~Mintun, J.~Pan, K.~V. Alwala, N.~Carion, C.-Y.
  Wu, R.~Girshick, P.~Doll{\'a}r, and C.~Feichtenhofer, ``Sam 2: Segment
  anything in images and videos,'' \emph{arXiv preprint arXiv:2408.00714},
  2024.

\bibitem{chen2024subobject}
D.~Chen, S.~Cahyawijaya, J.~Liu, B.~Wang, and P.~Fung, ``Subobject-level image
  tokenization,'' \emph{arXiv preprint arXiv:2402.14327}, 2024.

\bibitem{zou2024segment}
X.~Zou, J.~Yang, H.~Zhang, F.~Li, L.~Li, J.~Wang, L.~Wang, J.~Gao, and Y.~J.
  Lee, ``Segment everything everywhere all at once,'' \emph{Advances in Neural
  Information Processing Systems}, vol.~36, 2024.

\bibitem{nauta2023anecdotal}
M.~Nauta, J.~Trienes, S.~Pathak, E.~Nguyen, M.~Peters, Y.~Schmitt,
  J.~Schl{\"o}tterer, M.~Van~Keulen, and C.~Seifert, ``From anecdotal evidence
  to quantitative evaluation methods: A systematic review on evaluating
  explainable ai,'' \emph{ACM CSUR}, vol.~55, no. 13s, pp. 1--42, 2023.

\bibitem{zhou2021evaluating}
J.~Zhou, A.~H. Gandomi, F.~Chen, and A.~Holzinger, ``Evaluating the quality of
  machine learning explanations: A survey on methods and metrics,''
  \emph{Electronics}, vol.~10, no.~5, p. 593, 2021.

\bibitem{nauta2023co}
M.~Nauta and C.~Seifert, ``The co-12 recipe for evaluating interpretable
  part-prototype image classifiers,'' in \emph{Proc. World Conference on
  Explainable Artificial Intelligence}.\hskip 1em plus 0.5em minus 0.4em\relax
  Springer, 2023, pp. 397--420.

\bibitem{schlinge2025comprehensiveevaluationprototypeneural}
P.~Schlinge, S.~Meinert, and M.~Atzmueller, ``Comprehensive evaluation of
  prototype neural networks,'' \emph{arXiv preprint arXiv:2507.06819}, 2025.

\bibitem{huang2023evaluation}
Q.~Huang, M.~Xue, W.~Huang, H.~Zhang, J.~Song, Y.~Jing, and M.~Song,
  ``Evaluation and improvement of interpretability for self-explainable
  part-prototype networks,'' in \emph{Proc. IEEE/CVF International Conference
  on Computer Vision}, 2023, pp. 2011--2020.

\bibitem{willard2024looks}
F.~Willard, L.~Moffett, E.~Mokel, J.~Donnelly, S.~Guo, J.~Yang, G.~Kim, A.~J.
  Barnett, and C.~Rudin, ``This looks better than that: Better interpretable
  models with protopnext,'' \emph{arXiv preprint arXiv:2406.14675}, 2024.

\bibitem{wang2021interpretable}
J.~Wang, H.~Liu, X.~Wang, and L.~Jing, ``Interpretable image recognition by
  constructing transparent embedding space,'' in \emph{Proc. IEEE/CVF
  international conference on computer vision}, 2021, pp. 895--904.

\bibitem{wah2011caltech}
C.~Wah, S.~Branson, P.~Welinder, P.~Perona, and S.~Belongie, ``The caltech-ucsd
  birds-200-2011 dataset,'' 2011.

\bibitem{khosla2011novel}
A.~Khosla, N.~Jayadevaprakash, B.~Yao, and F.-F. Li, ``Novel dataset for
  fine-grained image categorization: Stanford dogs,'' in \emph{Proc. CVPR
  workshop on fine-grained visual categorization}, vol.~2, no.~1, 2011.

\bibitem{krause20133d}
J.~Krause, M.~Stark, J.~Deng, and L.~Fei-Fei, ``3d object representations for
  fine-grained categorization,'' in \emph{Proc. IEEE International Conference
  on Computer Vision Workshops}, 2013, pp. 554--561.

\end{thebibliography}

\end{document}